\documentclass[10pt,twocolumn,letterpaper]{article}

\usepackage{iccv}
\usepackage{subcaption}
\usepackage{times}
\usepackage{graphicx}
\usepackage{amsmath}
\usepackage{amssymb}

\usepackage[table,xcdraw]{xcolor}
\usepackage{booktabs}
\usepackage{comment}
\usepackage{caption}
\usepackage{subcaption}
\usepackage{colortbl}

\usepackage{bm}
\usepackage{comment}
\usepackage{here}
\usepackage{url}

\usepackage{color}

\newcommand{\cRyo}[1]{{\color{black}{#1} } }  
\newcommand{\cRyoo}[1]{{\color{black}{#1} } }  
\newcommand{\cRio}[1]{{\color{black}{#1}}}
\newcommand{\cHiro}[1]{{\color{black}{#1}}}

\newcommand{\cSora}[1]{{\color{black}{#1}}}

\makeatletter
\def\blfootnote{\gdef\@thefnmark{}\@footnotetext}
\makeatother

\iccvfinalcopy % *** Uncomment this line for the final submission

\def\iccvPaperID{****} % *** Enter the ICCV Paper ID here

\ificcvfinal\fi

\def\tableA{

\begin{table*}[t]
    \centering
    \caption{Comparison of pre-training methods. Best values at each dataset scale are in bold. $^\diamondsuit$ indicates a subset that consists of one randomly sampled image per class. ViT-T is used for all experiments. SL: supervised learning using cross-entropy loss, SSL: self-supervised learning using DINO \cite{CaronICCV2021_dino}. Fine-tuning accuracies are reported.}
    \vspace{-5pt}
    \scalebox{0.9}{
    \begin{tabular}{lccccccccc|c} \toprule[0.8pt]
        Pre-training & \#Img & Type & C10 & C100 & Cars & Flowers & VOC12 & P30 & IN100 & Average \\\midrule[0.5pt]
        Scratch & -- & -- &  78.3 & 57.7 & 11.6 & 77.1 & 64.8 & 75.7 & 73.2 & 62.6 \\
        \midrule[0.5pt]
        Places-365 \cite{ZhouTPAMI2017_Places} & 1.80M & SL & 97.6 & 83.9 & 89.2 & 99.3 & 84.6 & -- & \textbf{89.4} & -- \\
        ImageNet-1k \cite{DengCVPR2009_ImageNet} & 1.28M & SL & \textbf{98.0} & \textbf{85.5} & \textbf{89.9} & \textbf{99.4} & \textbf{88.7} & \textbf{80.0} & -- & --  \\
        ImageNet-1k \cite{DengCVPR2009_ImageNet} & 1.28M & SSL & 97.7 & 82.4 & 88.0 & 98.5 & 74.7 & 78.4 & 89.0 & 86.9 \\
        PASS~\cite{AsanoNeurIPS2021_pass}& 1.43M & SSL & 97.5 & 84.0 & 86.4 & 98.6 & 82.9 & 79.0 & 82.9 & \textbf{87.8} \\
        FractalDB-1k~\cite{KataokaACCV2020} & 1.00M & FDSL & 96.8 & 81.6 & 86.0 & 98.3 & 80.6 & 78.4 & 88.3 & 87.1 \\
        RCDB-1k~\cite{Kataoka_2022_CVPR} & 1.00M & FDSL & 97.0 & 82.2 & 86.5 & 98.9 & 80.9 & 79.7 & 88.5 & 87.6 \\
        \midrule[0.5pt]
        ImageNet-1k{$^\diamondsuit$} & \textbf{1,000} & SL & 94.3 & 76.9 & 57.3 & 94.8 & 73.8 & 78.2 & 84.3 & 79.9 \\
        ImageNet-1k{$^\diamondsuit$} & \textbf{1,000} & SSL & 94.9 & 78.0 & 71.2 & 94.6 & 75.5 & 78.6 & 84.9 & 82.5 \\
        2D-OFDB-1k (ours) & \textbf{1,000} & FDSL & 96.9 & 84.0 & 84.5 & 97.1 & 79.9 & 79.9 & 88.0 & 87.2 \\
        \rowcolor[gray]{0.8} 2D-OFDB-1k w/ Aug. (ours) & \textbf{1,000} & FDSL & \textbf{97.2} & \textbf{85.3} & \textbf{87.6} & 98.3 & \textbf{81.4} & \textbf{80.4} & \textbf{89.5} & \textbf{88.5} \\
        3D-OFDB-1k (ours) & \textbf{1,000} & FDSL & 97.1 & 83.8 & 85.5 & \textbf{98.4} & 80.8 & 80.0 & 89.1 & 87.8 \\    
        \rowcolor[gray]{0.8} 3D-OFDB-1k w/ Aug. (ours) & \textbf{1,000} & FDSL & 97.0 & 84.7 & 85.6 & 98.3 & 81.2 & 79.8 & 88.9 & 87.9 \\
        \bottomrule[0.8pt]
    \end{tabular}
    }
    \vspace{-10pt}
    \label{tab:comparison}
\end{table*}

}

\def\tableimagenet{
	\caption{Scaled models/datasets with combinations of 21k categories and ViT-T/B models are used in ImageNet-1k fine-tuning. \cHiro{We also list GPU hours, batch size (`Batch'), and number of iterations (`\#Iterations') in ViT-B pre-training.} }
 \vspace{-5pt}
    \scalebox{0.9}{
    \begin{tabular}{lcccc|ccc} \toprule[0.8pt]
        Pre-training & \#Img & Type & ViT-T & ViT-B & \cHiro{GPU hours} & \cHiro{Batch} & \cHiro{\#Iterations} \\
        \midrule[0.5pt]
        Scratch & -- & -- & 72.6 & 79.8 & -- & \cHiro{--} & \cHiro{--} \\
        \midrule[0.5pt]
        ImageNet-21k & 14M & SL & \textbf{74.1} & 81.8 & 3,657 & \cHiro{8,192} & \cHiro{300k} \\
        FractalDB-21k & 21M & FDSL & 73.0 & 81.8 & 5,120 & \cHiro{8,192} & \cHiro{300k} \\
        ExFractalDB-21k & 21M & FDSL & 73.6 & \textbf{82.7} & 5,120 & \cHiro{8,192} & \cHiro{300k}  \\
        RCDB-21k & 21M & FDSL & 73.1 & 82.4 & 5,120 & \cHiro{8,192} & \cHiro{300k} \\
        \midrule[0.5pt]
        ImageNet-21k{$^\diamondsuit$} & 21k & SL & 71.0 & 81.1 & 1,132 & \cHiro{1,024} & \cHiro{300k}  \\
        \rowcolor[gray]{0.8} 2D-OFDB-21k & 21k & FDSL & \textbf{73.8} & 82.2 & 1,088 & \cHiro{1,024} & \cHiro{300k} \\
        \rowcolor[gray]{0.8} 3D-OFDB-21k & 21k & FDSL & 73.7 & \textbf{82.7} & 1,088 & \cHiro{1,024} & \cHiro{300k} \\
        \bottomrule[0.8pt]
    \end{tabular}
    }
    \vspace{-15pt}
    \label{tab:comparison_imagenet1k}
}

\def\tableB{
\begin{table*}[t]
\centering
\tableimagenet
\end{table*}
}

\def\tableG{
\begin{table}[t]
\tablecoco
\end{table}
}
\def\tableC{
\begin{table}[t]
    \centering
    \caption{Exploratory experiment on one-instance ImageNet pre-training. Preprocessing ImageNet-1k$^\diamondsuit$ (1-instance) dataset with RGB images, we convert \{gray, binary, canny\} images for pre-training for each dataset. In this experiment, ImageNet-1k$^\diamondsuit$ (Canny) achieved the best accuracy \cRyo{in ImageNet-1k$^\diamondsuit$ sets \{RGB, gray, binary, canny\}}.}
    \vspace{-5pt}
    \scalebox{0.85}{
    \begin{tabular}{llcccc} 
\toprule[0.8pt]
        Pre-training & C10 & C100 & IN100 & P30 \\\midrule
        ImageNet-1k$^\diamondsuit$ & 94.3 & 76.9 & 84.3 & 78.2 \\ 
         -- Gray scale~\cite{gray_scale}  & 96.1 & 81.2 & 87.8 & 79.9 \\ 
         -- Binary~\cite{oths_method} & 96.7 & 82.7 & 88.8 & 79.9 \\ 
         -- Canny~\cite{canny} & 96.5 & 82.8 & 87.7 & 80.3 \\
\midrule
        \rowcolor[gray]{0.8} 2D-OFDB-1k & 96.9 & \textbf{84.0} & 88.0 & \textbf{80.4} \\
        \rowcolor[gray]{0.8} 3D-OFDB-1k & \textbf{97.1} & 83.8 & \textbf{89.1} & 80.0 \\ \bottomrule[0.8pt]
    \end{tabular}
    }
    \label{tab:1ins_oin}
    \vspace{-5pt}
\end{table}
}

\def\tablecoco{
\caption{Comparison of object detection and instance segmentation. Several pre-trained models were validated on the COCO dataset. The best values for each learning type are shown in bold.}
	\vspace{-5pt}
 \centering
 \scalebox{0.9}{
    \begin{tabular}{lccc} \toprule[0.8pt]
        \hspace{-3pt}Pre-training &  COCO Det & COCO Inst Seg \\
         &  AP$_{50}$ / AP / AP$_{75}$ & AP$_{50}$ / AP / AP$_{75}$ \\
        \midrule[0.5pt]
        \hspace{-3pt}Scratch  & 63.7 / 42.2 / 46.1 & 60.7 / 38.5 / 41.3 \\
        \midrule[0.5pt]
         \hspace{-3pt}ImageNet-1k  & 69.2 / 48.2 / 53.0 & 66.6 / 43.1 / 46.5 \\
        \hspace{-3pt}ImageNet-21k & \textbf{70.7} / \textbf{48.8} / \textbf{53.2} & \textbf{67.7} / \textbf{43.6} / \textbf{47.0} \\
        \hspace{-3pt}ExFractalDB-1k\hspace{-16pt} & 69.1 / 48.0 / 52.8 & 66.3 / 42.8 / 45.9 \\
        \hspace{-3pt}ExFractalDB-21k\hspace{-16pt} & 69.2 / 48.0 / 52.6 & 66.4 / 42.8 / 46.1 \\
        \hspace{-3pt}RCDB-1k & 68.3 / 47.4 / 51.9 & 65.7 / 42.2 / 45.5 \\
        \hspace{-3pt}RCDB-21k & 67.7 / 46.6 / 51.2 & 64.8 / 41.6 / 44.7 \\
        \midrule[0.5pt]
        \hspace{-3pt}ImageNet-\cSora{2}1k{$^\diamondsuit$} & \cSora{63.8} / \cSora{42.0} / \cSora{45.5} & \cSora{60.7} / \cSora{38.3} / \cSora{41.0} \\
         \hspace{-3pt}2D-OFDB-21k &
         \cSora{\textbf{67.6}} / \cSora{\textbf{46.4}} / \cSora{\textbf{51.0}} & \cSora{\textbf{64.6}} / \cSora{\textbf{41.6}} / \cSora{\textbf{44.7}} \\
        \hspace{-3pt}3D-OFDB-21k & \cSora{67.1} / \cSora{46.3} / \cSora{51.0} & \cSora{64.4} / \cSora{41.4} / \cSora{44.4} \\
        \bottomrule[0.8pt]
    \end{tabular}
    }
    \vspace{-7pt}
    \label{tab:comparison_detection_segmentation}

}
\def\tableD{
\begin{table*}[t]
    \centering
    \caption{Pre-training with small datasets. The pre-training and fine-tuning setting in \cite{ViT2040} is used for evaluation. \cHiro{Note that the values `2,040 -- 8,144' correspond to the number of pre-training images at each dataset. \textbf{\underline{Best}} and \textbf{second-best} scores are in underlined bold
and bold, respectively.}}
 \vspace{-5pt}
 \scalebox{0.9}{
    \begin{tabular}{lcccccccc|c} 
    \toprule[0.8pt]
        Pre-training & \cHiro{\#Img} & Flowers & Pets & DTD & Indoor-67 & CUB & Aircraft & Cars & Average \\
        \midrule[0.5pt]
        Scratch & -- & 76.4 & 67.2 & 44.2 & 58.7 & 54.4 & 23.0 & 78.6 & 57.5\\
        SimCLR~\cite{ChenICML2020} & 2,040 -- \cHiro{8,144} & 90.1 & 82.8 & 62.3 & 66.6 & \textbf{68.5} & 74.4 & 89.3 & 76.3\\
        IDMM~\cite{ViT2040} & 2,040 -- \cHiro{8,144} & 92.4 & 83.2 & 66.9 & 68.5 & \textbf{\underline{69.8}} & 73.4 & 87.8 & 77.4\\
        IDMM-ImageNet~\cite{ViT2040} & 2,040 & 90.5 & 82.4 & 66.8 & {\bf \underline{68.8}} & 66.8 & 91.8 & 87.6 & 79.2\\
        \rowcolor[gray]{0.8} 2D-OFDB-1k (ours) & 1,000 & \textbf{\underline{93.7}} & \textbf{\underline{84.6}} & \textbf{\underline{67.5}} & 66.1 & 67.7 & \textbf{\underline{95.0}} & \textbf{\underline{91.0}} & \textbf{80.8}\\
        \rowcolor[gray]{0.8} 3D-OFDB-1k (ours) & 1,000 & \textbf{92.8} & \textbf{\underline{84.6}} & \textbf{\underline{67.5}} & \textbf{68.6} & 67.9 & \textbf{94.6} & \textbf{90.4} & \textbf{\underline{80.9}} \\
        \bottomrule[0.8pt] 
    \end{tabular}
    }
    \vspace{-15pt}
    \label{tab:comparisons_vit_pretraining_limited_data}
\end{table*}
}

\def\tableH{
\begin{table}[t]
    \centering
    \caption{Analysis on 3D-OFDB. Although ExFractalDB-1k previously adjusted one-axis with yaw angle, we adjusted three axes with \{roll, pitch, yaw\} angles.}
    \vspace{-5pt}
    \scalebox{0.9}{
    \begin{tabular}{llc} \toprule[0.8pt]
        Pre-Training & Axis & Acc. \\\midrule[0.5pt]
        3D-OFDB & 1 \cHiro{(yaw)} & \textbf{83.8} \\
        & 2 \cHiro{(pitch, yaw)} & 82.8 \\
        & 3 \cHiro{(roll, pitch, yaw)} & 82.7 
        \\\midrule[0.5pt]
        ExFractalDB & \cHiro{3 (roll, pitch, yaw)} & 83.1 
        \\\bottomrule[0.8pt]
    \end{tabular}
    }
    \vspace{-7pt}
    \label{tab:oefdb_viewpoint}
\end{table}
}

\def\tableI{
\begin{table}[t]
    \centering
    \caption{Effects of data augmentation for fractal images. Data-efficient image Transformer (DeiT) augmentation setting is used as a default. 
    }
    \vspace{-5pt}
    \scalebox{0.9}{
    \begin{tabular}{l|cccccc} 
    \toprule[0.8pt]
    DeiT & \checkmark & \checkmark & \checkmark & \checkmark & \checkmark & \checkmark \\
    IFS & & \checkmark & & &\\
    Rotation & & & \checkmark & & &\\
    Rand. Pat. & & &  & \checkmark & & \checkmark \\
    Rand. Text. & & &  &  &  \checkmark & \checkmark \\
    \midrule[0.5pt]
    2D-OFDB & 84.0 & 81.6 & 84.1 & \textbf{85.3} & 84.8 & 84.3\\
    3D-OFDB & 83.8 & - & - & 84.7 & \textbf{85.1} & \textbf{85.1} \\ \toprule[0.8pt]
    \end{tabular}
    }
    \vspace{-13pt}
    \label{tab:ofdb_dataaug}
\end{table}
}

\def\tableJ{

\begin{table}[t]
    \centering
    
    \caption{Datasets rendering time by full-instance (FractalDB/ExFractalDB-1k) and one-instance (2D/3D-OFDB-1k). We separately show fractal category search (Search), image rendering (Render), and total time (Total). The values are given in hours.}
    \vspace{-5pt}
    \scalebox{0.9}{
    \begin{tabular}{lccc} \toprule[0.8pt]
    Dataset & Search & Render & Total \\
    \midrule[0.5pt]
    FractalDB-1k & 2.37 & 16.86 & 19.23 \\
    2D-OFDB-1k & 2.37 & 0.41 & 2.78 \\ \toprule[0.5pt]
    ExFractalDB-1k & 0.53 & 5.43 & 5.96  \\
    3D-OFDB-1k & 0.53 & 0.04 & 0.57  \\ \toprule[0.8pt]
    \end{tabular}
    }
    \vspace{-5pt}
    \label{tab:rendering_time}
    \vspace{-5pt}
\end{table}
}

\begin{document}

%%%%%%%%% TITLE
\title{
Pre-training Vision Transformers with Very Limited Synthesized Images
}
\vspace{-20pt}
\author{Ryo Nakamura$^{1,2,*}$, Hirokatsu Kataoka$^{1,*}$, Sora Takashima$^{1,3}$,\\ Edgar Josafat Martinez Noriega$^{1,3}$, Rio Yokota$^{1,3}$, Nakamasa Inoue$^{1,3}$\\
$^{1}$National Institute of Advanced Industrial Science and Technology (AIST),\\
$^{2}$Fukuoka University, 
$^{3}$Tokyo Institute of Technology  \\
% \footnotesize{\url{https://github.com/ryoo-nakamura/OFDB/}}
}

\maketitle
\ificcvfinal\thispagestyle{empty}\fi
\vspace{-20pt}

%%%%%%%%% ABSTRACT
\begin{abstract}

\cRio{Formula-driven supervised learning (FDSL) is a pre-training method that relies on synthetic images generated from mathematical formulae such as fractals.
Prior work on FDSL has shown that pre-training vision transformers on such synthetic datasets can yield competitive accuracy on a wide range of downstream tasks.
These synthetic images are categorized according to the parameters in the mathematical formula that generate them.
In the present work, we hypothesize that the process for generating different instances for the same category in FDSL, can be viewed as a form of data augmentation.
We validate this hypothesis by replacing the instances with data augmentation, which means we only need a single image per category.
Our experiments shows that this one-instance fractal database (OFDB) performs better than the original dataset where instances were explicitly generated.
We further scale up OFDB to 21,000 categories and show that it matches, or even surpasses, the model pre-trained on ImageNet-21k in ImageNet-1k fine-tuning.
The number of images in OFDB is 21k, whereas ImageNet-21k has 14M.
This opens new possibilities for pre-training vision transformers with much smaller datasets.
}

\end{abstract}

%%%%%%%%% BODY TEXT
\vspace{-15pt}
\section{Introduction}

\blfootnote{\hspace{-16pt}
$^{*}$equal contribution 
}
\blfootnote{\hspace{-16pt}
GitHub code : \url{https://github.com/ryoo-nakamura/OFDB/}
}

\label{sec:intro}
% \begin{figure}[t]
%   \centering
%   \includegraphics[width=1.0\linewidth]{figure/figA.pdf}
%   \vspace{-18pt}
%   \caption{(a) Fine-tuning accuracy on CIFAR-100 with respect to the number of images for pre-training. The line plot for ImageNet-1k indicates the results when using random sampling for reducing the amount of data for pre-training. (b) One-instance fractal database (OFDB), which consists of only 1,000 images.}
%   \vspace{-12pt}
%   %   SL（ImageNet-1k）/FDSL（FractalDB-1k）とインスタンス数の関係性。
%   \label{fig:acc_ins_tradeoff}
% \end{figure}

\begin{figure}[t]
  \centering
  \scalebox{0.93}{
  \includegraphics[width=0.97\linewidth]{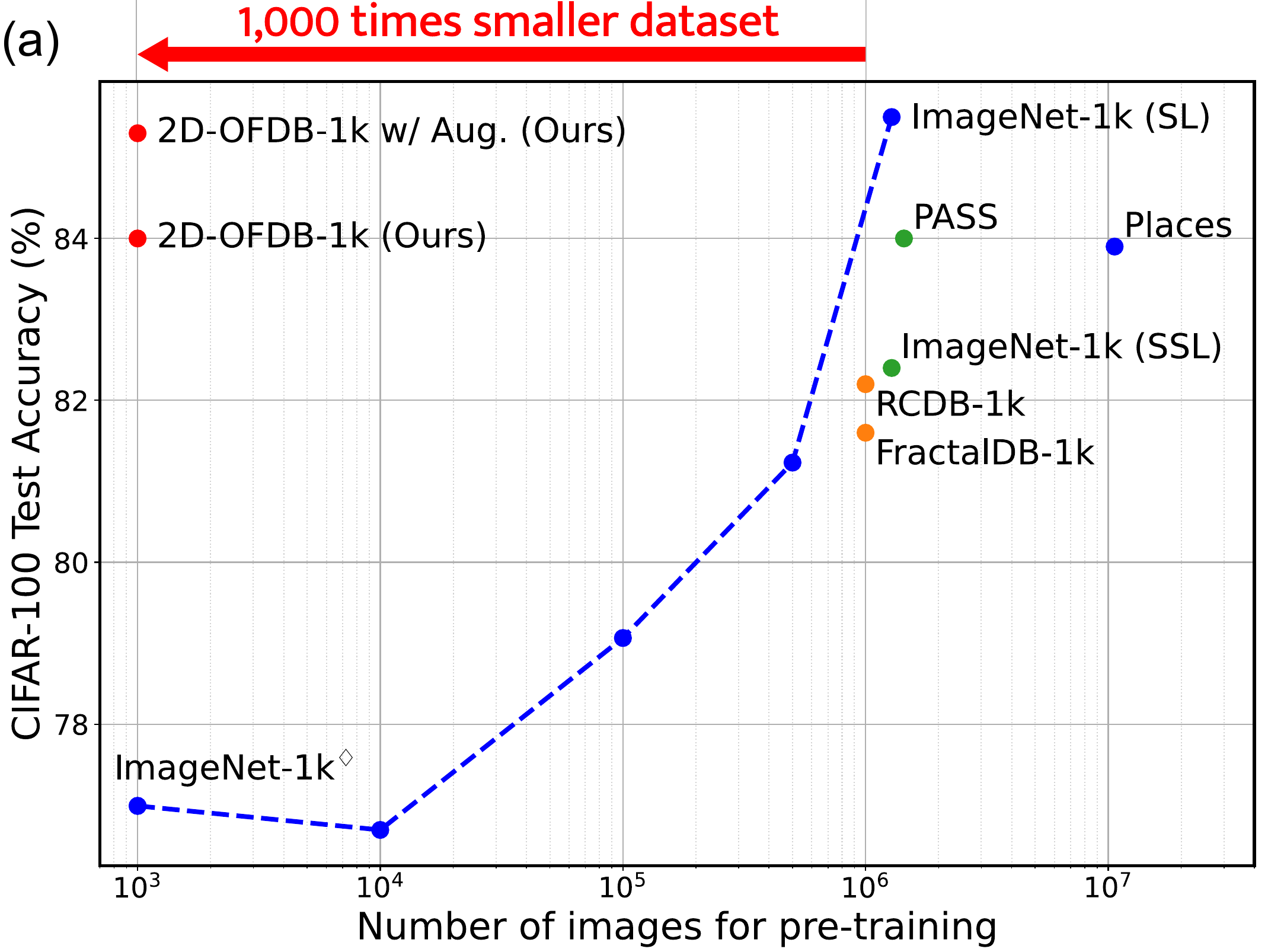}
  \label{fig:acc_ins_tradeoff-a}
  }
  \\
\scalebox{0.95}{
  \includegraphics[width=0.97\linewidth]{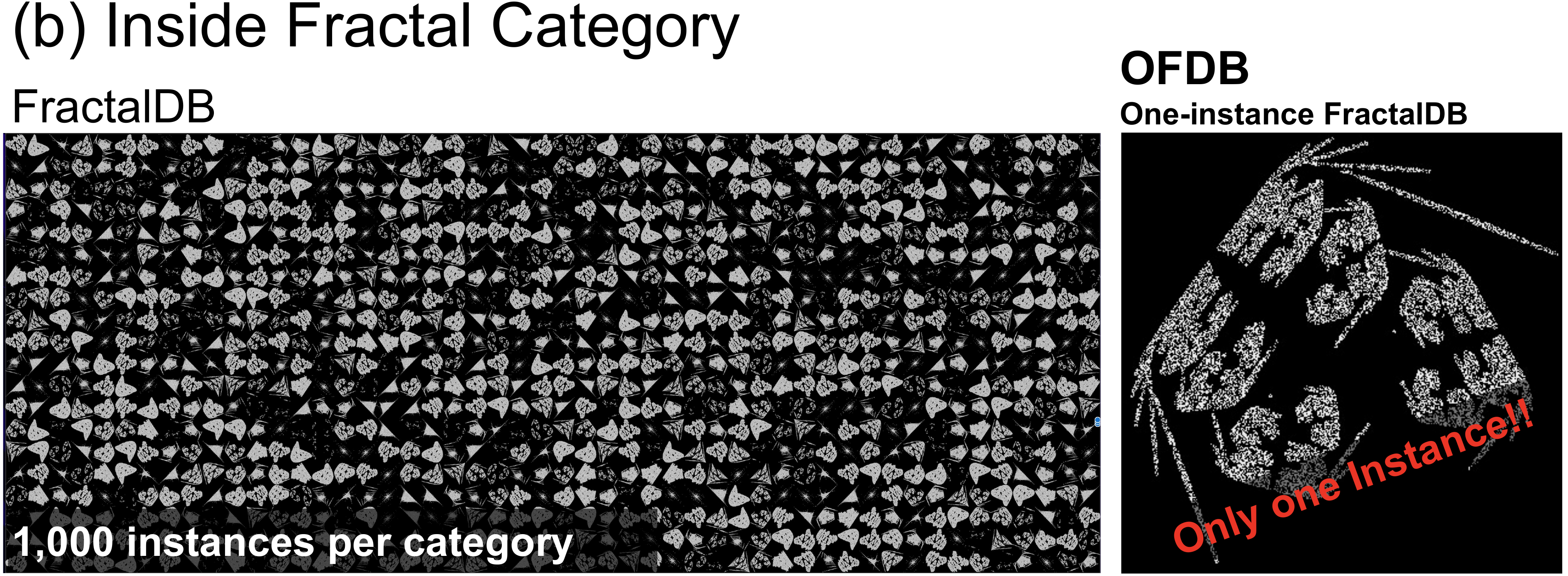}
  \label{fig:acc_ins_tradeoff-b}
  }
  \vspace{-5pt}
  \caption{
  \cRyoo{(a) Fine-tuning accuracy on CIFAR-100 for the number of images for pre-training. The line plot for ImageNet-1k indicates the results when using random sampling to reduce the data for pre-training. (b) One-instance fractal database (OFDB) consists of only 1,000 images in total. The figure shows the category representation. OFDB contains a single instance per category.}
  }
\label{fig:acc_ins_tradeoff}
\vspace{-5pt}
\end{figure}

% Pre-training has become a standard procedure when training deep neural networks in computer vision~\cite{PrabhuWACV2021_imagedatasets,ZhouTPAMI2017_Places}.
% Pre-trained models are known to exhibit superior convergence and generalization for downstream tasks, as compared to models that are trained from scratch. 
% However, pre-training large vision models requires an enormous amount of data, so pre-training state-of-the-art vision models is very expensive, both in terms of the required amount of data and computation. 
Pre-training has become a standard procedure when training deep neural networks in computer vision~\cite{PrabhuWACV2021_imagedatasets,ZhouTPAMI2017_Places}.
Pre-trained models are known to exhibit superior convergence and generalization for downstream tasks as compared to models that are trained from scratch. 
However, pre-training large vision models requires enormous data, which makes pre-training state-of-the-art vision models very expensive regarding the required amount of data and computation. 

%the required amount of computation.
During the past decade, ImageNet has served as a common dataset for pre-training vision models~\cite{DengCVPR2009_ImageNet}.
Models pre-trained on the classification task of 1.28M images in ImageNet were transferred to Object Detection~\cite{EveringhamIJCV2015_voc,LinECCV2014_coco,ZhouTPAMI2017_Places}, Semantic Segmentation~\cite{segformer,Ranftl_2021_ICCV}, and Video Recognition~\cite{Kinetics, vivit}. %tasks.
%This served an important role in the permeation of deep learning in computer vision.
In the 2020s, the pre-training of Vision Transformers (ViT)~\cite{DosovitskiyICLR2021} has become increasingly popular.
However large vision transformers are said to require datasets that are much larger than ImageNet, such as JFT-300M/3B~\cite{SunICCV2017_jft300m}, in order to achieve their true potential.

\vspace{-3pt}
Pre-training increasingly larger models on increasingly larger datasets has shown a monotonic improvement in the accuracy of downstream tasks.
However, creating datasets with billions of labeled images is not the ultimate solution to all our problems.
First, the cost of manually  labeling such huge datasets is prohibitive.
Self-supervised learning (SSL) has received great attention due to its competitive performance when used to pre-train large models without requiring labeled images.
\cHiro{DINO}~\cite{CaronICCV2021_dino}, MoCoV3~\cite{ChenICCV2021_mocov3}, and 
\cHiro{BEiT}~\cite{bao2022beit} have shown promising results in this regard.
However, SSL still relies on a vast amount of unlabeled data.

Therefore, we should pause and ask the question, ``Are we using all this data efficiently?"
There have been efforts in this direction to reduce the amount of pre-training data while retaining accuracy on downstream tasks.
For ViTs, 
\cHiro{DeiT}~\cite{TouvronICML2021} has demonstrated that distillation and data augmentation can enhance the pre-training effect of ImageNet-1k to match that of larger datasets.
\cHiro{MAE}~\cite{MaskedAutoencoders2021} also show a high performance, even when trained on ImageNet-1k.
More recent studies, such as ``Training Vision Transformers with Only 2,040 Images"~\cite{ViT2040}, show how ViTs can be pre-trained on \cHiro{relatively} small datasets.
% Both approaches rely on internal (within the model) and external (data) augmentation of their data in order to extract as much information as possible from each image.

% Another approach that is orthogonal to the those above is the pre-training of a ViT using formula-driven supervised learning (FDSL)~\cite{NakashimaarXiv2021, KataokaACCV2020, Kataoka_2022_CVPR, PerlinToG2002, DosovitskiyICLR2021}, in which not even a real image is required.
\cRyoo{Another approach to those above is pre-training} a ViT using formula-driven supervised learning (FDSL)~\cite{NakashimaarXiv2021, KataokaACCV2020, Kataoka_2022_CVPR, PerlinToG2002, DosovitskiyICLR2021}, in which not even a real image is required.
In a follow-up study, Kataoka \textit{et al.}~\cite{Kataoka_2022_CVPR} created an alternative synthetic dataset with emphasis on the contours in the image and showed that it is possible to surpass the accuracy of a ViT pre-trained on ImageNet-21k by using a synthetic dataset of the same size.
\cHiro{FDSL method} can generate the labels automatically from the parameters used to generate the images, so there is no labeling cost.
Furthermore, unlike \cHiro{SSL}, FDSL does not even require real images.
The fact that a ViT pre-trained on synthetic datasets can outperform a ViT pre-trained on a fairly large human-labeled dataset ImageNet-21k is significant. 
However, synthesizing more than a million images is costly.
It is possible that many of the images in these synthetic datasets are redundant or are simply not contributing to the pre-training \cHiro{since these synthetic images are expanded with basic procedures from a single image. In case of FractalDB~\cite{KataokaACCV2020}, a single image inside of category was augmented to 1,000 instances with image rotation, parameter fluctuation, and patch patterns. It is natural to rely on the combination of data augmentation methods in pre-training phase. In this context, it may be possible to reach the same fine-tuning accuracy for both one instance with the same data augmentations and preprocessed 1,000 instances. }

In this paper, we present an FDSL approach to pre-train a ViT with a single instance per category. Therefore, we require only 1,000 images when the dataset contains 1,000 categories.
The proposed dataset, i.e., the one-instance fractal database (OFDB), significantly improves data efficiency under the assumption that data augmentation is used during pre-training.
In previous FDSL datasets, the different instances for each category are created through some form of manipulation of the original image that defines that category.
Therefore, it is quite natural to wonder whether these instances can be created through data augmentation techniques.
Although a detailed description is provided in Section~\ref{sec:experiments}, we disclose that the fractal instances can be replaced by data augmentation, e.g., image rotation. Based on the above considerations, we use a novel FDSL dataset that only requires a single image per category and training with data augmentation including random pattern augmentation and random texture augmentation as proposed data augmentation methods for FDSL pre-training.
We validate this hypothesis by creating small, yet effective, pre-training datasets, 2D-OFDB and 3D-OFDB, respectively. 

Our main contributions can be summarized as follows:

\noindent \cHiro{\textbf{\underline{Conceptual contribution.}}} We propose two datasets, 2D-OFDB and 3D-OFDB, which consist of only one representative fractal per category.
For example, 2D-OFDB-1k consists of only 1,000 images but enables \cHiro{ViT to effectively} learn visual representations for image classification. \cHiro{Along this line, we also implement random pattern augmentation and random texture augmentation for fractal pre-training.} 

\noindent \cHiro{\textbf{\underline{Experimental contribution.}}} We show that OFDBs achieve comparable performance to well-defined million-scale datasets \cHiro{(Figure~\ref{fig:acc_ins_tradeoff} and Table~\ref{tab:comparison})}. Furthermore, we show that the computational \cRyoo{time} of pre-training is reduced by \cHiro{78.7\%}. \cHiro{In ImageNet-1k fine-tuning, 2D/3D-OFDB-21k performed at equal or better rates than baseline pre-training datasets with only 21k images. (see Table~\ref{tab:comparison_imagenet1k}).} We also show that OFDBs perform better than state-of-the-art methods for training ViTs on small datasets~\cite{ViT2040} \cHiro{(Table~\ref{tab:comparisons_vit_pretraining_limited_data})}. 
\vspace{-5pt}
\section{Related Work}

\subsection{Formula-driven supervised learning \cHiro{(FDSL)}}

\cHiro{FDSL} is a form of \cHiro{learning strategy} in which images and their corresponding labels are generated from a mathematical formula~\cite{KataokaACCV2020,KataokaICCV2021WS,NakashimaarXiv2021,Kataoka_2022_CVPR,PerlinToG2002,AndersonWACV2022}.
Training on such synthetic datasets frees us from various ethical issues, e.g., societal biases and handling of copyrights and personal information~\cite{YangFAT2020,AsanoNeurIPS2021_pass}.
One of the representative datasets for FDSL is FractalDB, which generates fractal images from an iterated function system (IFS)~\cite{KataokaACCV2020}.
% FractalDB does not require manual labeling of the data, nor does it require users to download real images.

Nakashima \textit{et al.}~\cite{NakashimaarXiv2021} demonstrated that a ViT can be successfully pre-trained on FractalDB, which results in competitive accuracy on downstream tasks to a ViT pre-trained on ImageNet.
% Anderson \textit{et al.}~\cite{AndersonWACV2022} showed that the pre-training can be further improved by using SVD to search for the optimal IFS parameters.
% Baradad \textit{et al.}~\cite{NEURIPS2021_14f2ebea} used a different synthetic dataset, Deadleaves, and demonstrated that it is possible to learn visual representations from such datasets.
More recently, Kataoka \textit{et al.}~\cite{Kataoka_2022_CVPR} extended the FractalDB to two other datasets (ExFractalDB and RCDB), which comprise images with more emphasis on contours rather than textures.
% These datasets pre-train ViT on ImageNet-21k and compare images with ExFractalDB-21k and RCDB-21k, which are synthetic datasets with the same number of categories and roughly the same number of images.
When fine-tuned on ImageNet-1k, the accuracy of the synthetic datasets (ExFractalDB-21k and RCDB-21k) exceeds that of ImageNet-21k.
This is a significant result that raises fundamental questions regarding the role of \cHiro{real} images when pre-training ViTs.
Furthermore, there is ample room for improvement regarding the quality of these synthetic datasets.
The ExFractalDB-21k and RCDB-21k dataset each have 21k categories and 1k instances per category.
The instances are created by manipulating the original image, using a process similar to data augmentation.
In the present study, we consider the possibility of replacing these instances with standard \cHiro{or our proposed} data augmentation techniques in image classification.%~\cite{TouvronICML2021}.

\vspace{-5pt}
\subsection{Pre-training ViT on Limited Data}
\vspace{-3pt}
Since Dosovitskiy \textit{et al.} published the ViT~\cite{DosovitskiyICLR2021} paper in 2020, they have been replacing convolutional neural networks (CNN)~\cite{cnn,NIPS2012_c399862d} for various computer vision tasks.
However, large \cHiro{ViT} models require large datasets, such as ImageNet-21k (14M images) and JFT-300M (300M images), to reach their full potential.
The DeiT~\cite{TouvronICML2021} uses data augmentation and distillation from CNNs to achieve the same pre-training effect using only ImageNet-1k (1.28M images).
\cHiro{SSL}, such as that by DINO~\cite{CaronICCV2021_dino}, MoCoV3~\cite{ChenICCV2021_mocov3}, BeiT~\cite{bao2022beit}, and MAE~\cite{MaskedAutoencoders2021}, is also able to achieve similar performance using ImageNet-1k for pre-training.
Other studies have attempted to reduce the size of the pre-training dataset even further.
Cao \textit{et al.}~\cite{ViT2040} modified the structure of ViT in order to enhance the information extracted from images and were able to pre-train ViT using only 2,040 \cHiro{-- 8,144} images.
This is three orders of magnitude smaller than ImageNet-1k, which opens new possibilities for pre-training vision transformers on small datasets.

%In the present paper, 
\cHiro{We} propose a one-instance fractal database (OFDB), where each category has only one instance.
The hypothesis here is that the 1,000 instances of the original FractalDB can be replaced with standard \cHiro{or proposed} data augmentation techniques.
In this case, we only need a single instance per category, whereas the other instances are created during training through data augmentation.
%Prior research has shown that FractalDB can match the performance of ImageNet.
%Therefore, if our FractalDB with one instance per category can match the performance of the original FractalDB with 1,000 instances, we can effectively match the performance of ImageNet while using an extremely small number of images.

\def\figE{
\begin{figure}[t]
  \centering
  \includegraphics[width=0.93\linewidth]{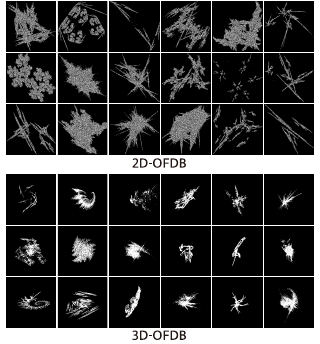}
  \vspace{-8pt}
  \caption{Example images of 2D/3D-OFDBs.}
  \label{figE}
\end{figure}
}

\def\figB{

\begin{figure}[t]
  \centering
\scalebox{0.87}{
  \subfloat[][Random patch augmentation.]{
  \includegraphics[width=0.85\linewidth]{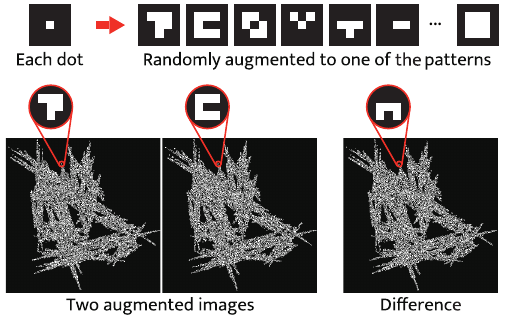}
%   \caption{\cRyo{Random patch augmentation}}
  \label{figB-a}
  }
  }
  \\
  \centering
  \scalebox{0.87}{
  \subfloat[][Random texture augmentation.]{
  \includegraphics[width=0.85\linewidth]{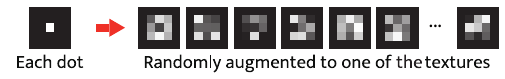}
  \label{figB-b}}
  }
  \vspace{-8pt}
  \caption{Proposed data augmentation methods.}
  
  \vspace{-5pt}
\end{figure}

}

\vspace{-5pt}
\section{Method}
\vspace{-3pt}

This section presents small and powerful datasets for FDSL.
In contrast to previous million-scale datasets, such as ImageNet-1k,
one of the proposed datasets, namely 2D-OFDB-1k, contains only 1,000 images for pre-training, but 2D-OFDB-1k enables \cHiro{ViT to effectively learn} visual representations for image classification. 
%In the following, we present the details of the proposed dataset.

%\subsection{なぜ，OFDBがmillion-scale datasetの事前学習効果に到達できるのか？？}
\cRyoo{
On the other hand, at the beginning of this section, we describe \textbf{`why fractal pre-training with one-instance per category can train visual representation'} as a curious scenario in FDSL datasets, especially on FractalDB. 
First, we consider instance augmentation of FractalDB. In FractalDB, the number of images is increased by (i) image rotation (x4), (ii) \cHiro{IFS} parameters fluctuation (x25), and (iii) patch patterns (x10) from \cHiro{a} representative image of the category found with category search. 
FractalDB~\cite{KataokaACCV2020} is pre-trained as a dataset of 1,000 instances of pre-processed images, but in a simple view, these could be replaced by data augmentation. Details will be discussed later, and Table~\ref{tab:ofdb_dataaug} shows that the accuracy is almost the same with and without Rotation (2D-OFDB w/o rotation 84.0 vs. w/ rotation 84.1 on CIFAR-100). \cHiro{On the other hand, the performance rates with IFS fluctuation are} significantly lower (2D-OFDB w/o IFS 84.0 vs. w/ IFS 81.6 on CIFAR-100). \cHiro{From these observations, we believe that a pre-training dataset with fewer instances like a dataset consists of one-instance per category} and using basic data augmentation are the key technologies to reduce training time to an equivalent or better accuracy level. As the result, \cHiro{an image augmentation with} patch patterns were newly \cHiro{implemented as random patch/texture augmentation} so that they could be implemented within data augmentation, and it became clear that accuracy could be improved while reducing the data size to \cHiro{0.1\% amount}.
% まず、FractalDBのインスタンス拡張について考える。FractalDBでは、カテゴリ探索により得られたカテゴリの代表画像から (i) image rotation（x4）、(ii) ifs parameters fructuation（x25）, (iii) patch patterns（x10）により増やされている。
% Original FractalDBではあらかじめ画像処理された1,000インスタンスのデータセットとして事前学習されているが、シンプルに考えてみると、これらはデータ拡張にて代替可能であるのではないだろうか。
% 詳細は後述するが、Table~\ref{tab:ofdb_dataaug}を見てみると、Rotationの有無ではほとんど精度が変わらない（2D-OFDB w/o rotation 84.0 vs. w/ rotation 84.1 on CIFAR-100）。さらに、IFSに至っては寧ろ精度を著しく下げる結果（2D-OFDB w/o IFS 84.0 vs. w/ IFS 81.6 on CIFAR-100）となっている。
% ここから、我々は one-instance によるデータセット構築と、事前学習効果を向上させる基礎的なデータ拡張のみを用いることこそが学習時間を削減し、同等以上の精度向上を実現するキーテクノロジーであると信じる。
% 結果的に、patch patterns はデータ拡張内で実装できるようRandom Patch/Texture Augmentationとして新規に提案、IFSを除外することでデータセットサイズを1/1,000にしつつも精度向上できることが明らかとなった。
}

\vspace{-5pt}
\subsection{Problem Settings}
\label{sec:setting}
\vspace{-3pt}

\noindent{\bf FDSL.}
The goal of FDSL is to pre-train neural networks without real images.
One of the most successful approaches
to achieve this goal is to synthesize a labeled dataset $D = \{(\bm{x}_{i}, \bm{y}_{i})\}_{i=1}^{N}$ based on some mathematical formulas, such as fractals ~\cite{KataokaACCV2020}, where $\bm{x}_{i}$ is a synthesized image, $\bm{y}_{i}$ is a one-hot label vector, and $N$ is the number of images.
For FDSL, the cross-entropy loss is used, which is given by
\begin{align} \label{eq:celoss}
\mathcal{L}_{\text{ce}}(\theta; D) = - \frac{1}{N} \sum_{i=1}^{N} \sum_{c=1}^{C} y_{i,c} \log p_{i,c},
\end{align}

where $\bm{p}_{i} = f_{\theta}(\bm{x}_{i}) \in \mathbb{R}^{C}$ is the output vector of a learnable network $f_{\theta}$, such as a \cHiro{ViT}, $\theta$ is a set of parameters, and $C$ is the number of categories.
Typically, the number of images $N$ should be equal to or more than one million in order to achieve good pre-training performance.

\noindent{\bf One-instance FDSL.}
The present paper proposes a challenging learning framework, namely {\it one-instance FDSL}, in an attempt at efficient and effective pre-training.
The goal of one-instance FDSL is to pre-train neural networks with a dataset of representative images $D = \{\bm{x}_{c}\}_{c=1}^{C}$, where $\bm{x}_{c}$ is a single image that represents category $c$, {\it i.e.,} dataset $D$ involves only one image instance per category. With this setting, the cross-entropy loss reduces to the following negative log-likelihood loss:
\begin{align}
\mathcal{L}_{\text{nl}} (\theta; D) = - \frac{1}{C} \sum_{c=1}^{C} \log p_{c,c}.
\end{align}
This setting dramatically improves data-efficiency of pre-training because the setting omits $N$ in Eq.~(\ref{eq:celoss}).
% However, this is a challenging problem because we have no intra-category variations.

\vspace{-5pt}
\subsection{One-instance Fractal Databases \cHiro{(OFDBs)}}

We propose \cHiro{OFDBs} involving representative images of fractals.
There are two variants, 2D-OFDB and 3D-OFDB, which consist of representative images of 2D and 3D fractals, respectively.

%\figE % Example images

\noindent{\bf 2D-OFDB.}
The first variant is a dataset that consists of representative 2D-fractal images.
In order to create fractals, the iterated function system (IFS) ~\cite{KataokaACCV2020} is used:$\operatorname{IFS}=\left\{\mathcal{X} ; w_{1}, w_{2}, \cdots, w_{M} ; p_{1}, p_{2}, \cdots, p_{M}\right\}$
% \begin{align}
% \operatorname{IFS}=\left\{\mathcal{X} ; w_{1}, w_{2}, \cdots, w_{M} ; p_{1}, p_{2}, \cdots, p_{M}\right\}
% \end{align}
where $\mathcal{X} = \mathbb{R}^{2}$ is a 2D Euclidean space, $w_{j} : \mathcal{X} \to \mathcal{X}$ is an affine transformation function, and $p_{j}$ is a probability.
Given an IFS and an initial point $v_{1} \in \mathcal{X}$, a fractal $S$ is obtained as a set of points $S = \{v_{t}\}_{t=1}^{\infty} \subset \mathcal{X}$ by $v_{t+1} = w^{*}(v_{t})$,
% \begin{align}
% v_{t+1} = w^{*}(v_{t}).
% \end{align}
where $w^{*}$ is a transformation sampled at each $t$ under the probability distribution $p(w^{*} = w_{j}) = p_{j}$.

The 2D-OFDB $D_{\text{2D}} = \{\bm{x}_{c}\}_{c=1}^{C}$ consists of $C$ representative images that are synthesized in the following three steps:
First, a set of iterated function systems $\{\text{IFS}_{c}\}_{c=1}^{C}$ is randomly sampled. We use the sampling algorithm proposed in ~\cite{KataokaACCV2020}.
Second, with each $\text{IFS}_{c}$, a fractal $S_{c}$ is randomly sampled. Finally, $S_{c}$ is rendered into $\bm{x}_{c}$.
Note that the original FractalDB ~\cite{KataokaACCV2020} samples 1,000 fractals from each $\text{IFS}_{c}$. However, we found that most fractals are redundant, especially with image rotation (see Table~\ref{tab:ofdb_dataaug} for details), if a data augmentation function is used when pre-training ViT.
For comparison with ImageNet, we create two datasets, \cHiro{2D-OFDB-1k/21k}, by setting $C=1,000$/$21,000$, respectively.

\noindent{\bf 3D-OFDB.} The second variant is a dataset consisting of 3D fractals.
%This dataset is the one-instance version of the ExFractlaDB~\cite{Kataoka_2022_CVPR}, which 
\cHiro{This dataset} uses the 3D Euclidean space $\mathcal{X} = \mathbb{R}^{3}$ and 3D affine transformations with IFSs.
The representative fractals are chosen by considering variance $\delta$ of point scattering in 3D space. We follow the previous study~\cite{Kataoka_2022_CVPR} to conduct the procedure. %in the same way as 2D-OFDB.
We create two datasets, \cHiro{3D-OFDB-1k/21k}, by setting $C=1,000$/$21,000$, respectively.
%%%Some example images are shown in Figure~\ref{figE}.

% \noindent{\bf なぜOFDBがうまく行くのか？？}

\tableA

\vspace{-5pt}
\subsection{Data augmentation for fractal images}
\vspace{-3pt}
%For training neural networks such as vision transformers, data augmentation is a necessary technique to improve the robustness against input perturbation.
In one-instance FDSL, we empirically found that the augmentation configuration proposed for the DeiT ~\cite{TouvronICML2021} is effective.
However, given representative fractal images, there is still room for improvement because the configuration of the DeiT is empirically optimized for pre-training with \cHiro{real} images.
Here, we present two additional augmentation functions that boost pre-training with OFDBs.
Also, note that our proposed data augmentation is OFDB-specific and cannot be applied to natural images.

\noindent{\bf Random pattern augmentation (Figure~\ref{figB-a}}).
The image $\bm{x}_{c}$ of the representative fractal $S_{c}$ is a binary (black-and-white) image of dots, each of which corresponds to a point $v_{t}$ of the fractal.
Given an image, $\bm{x}_{c}$, the random pattern augmentation augments each dot to a $3 \times 3$ pattern.
The patch patterns are randomly sampled from the uniform distribution over the set of all binary patterns (there are $2^{3 \times 3} = 512$ patterns).
% An example of two augmented images and the difference between these images are shown in Figure~\ref{figB-a}.
\cRyo{Figure~\ref{figB-a}} shows an example of two augmented images and the difference between the images.
We see that the overall fractal shape is the same for the two images, but the local patterns are different. Interestingly, the difference image obtained by this augmentation method makes the same fractal.

\noindent{\bf Random texture augmentation (Figure~\ref{figB-b}).}
The random texture augmentation augments each dot to a $3 \times 3$ gray-scale texture, 
% as shown in \cRyo{Figure~\ref{figB-b}}, 
where each pixel value is randomly sampled from the uniform distribution over $\{0, 1, \cdots, 255\}$.
Unlike random pattern augmentation, this augmentation makes dense images, where most of the nine pixels in each texture have non-zero values.

\figB % Augmentation methods

%-------------------------------------------------------------------------
\vspace{-7pt}
\section{Experiments}
\vspace{-3pt}
\label{sec:experiments}
%In this section, we evaluate the effects of one-instance FDSL using the proposed datasets, namely 2D-OFDB and 3D-OFDB. We also discuss the potential, limitation, and considerations based on the experimental results.

% Comparisons
\subsection{Comparison with State-of-the-art Datasets}
\vspace{-3pt}
\noindent \textbf{Fine-tuning results (Table~\ref{tab:comparison}).}
We conducted fine-tuning experiments on the
CIFAR-10 (C10) \cite{Krizhevsky2009_cifar}, CIFAR-100 (C100) \cite{Krizhevsky2009_cifar}, Cars~\cite{Krause3DRR2013_cars}, Flowers~\cite{Nilsback08_flowers}, ImageNet-100 (IN100)~\cite{KataokaACCV2020}, Places30 (P30)~\cite{KataokaACCV2020}, and Pascal VOC 2012 (VOC12)~\cite{EveringhamIJCV2015_voc} datasets.
The proposed OFDBs are compared with six pre-training datasets: ImageNet-1k 
% \cite{DengCVPR2009_ImageNet}
, Places-365
% ~\cite{ZhouTPAMI2017_Places}
, PASS
% ~\cite{AsanoNeurIPS2021_pass}
, FractalDB-1k
% ~\cite{KataokaACCV2020}
, RCDB-1k
% ~\cite{Kataoka_2022_CVPR}
, and ExFractalDB-1k
% ~\cite{Kataoka_2022_CVPR}
.
The results using a subset of ImageNet, which consists of 1,000 images obtained by randomly sampling one image per category, are also reported. Note that $^\diamondsuit$indicates the subset. This subset is the same size as OFDBs. Here, we assign the ViT-Tiny (ViT-T) model with standard DeiT training configurations, including hyper-parameters.

\tableB % Scaled models

\tableD % Small datasets

From the experimental results in Table~\ref{tab:comparison}, \cHiro{we see that OFDBs achieved a higher accuracy with million-scale FDSL datasets, such as ExFractalDB-1k.}
%we see that OFDBs achieved comparable performance with million-scale FDSL datasets, such as ExFractalDB-1k.
Although they did not always surpass \cHiro{SL and SSL} methods,
note that they often achieved similar performance rates with only 1,000 images for pre-training, which is approximately 0.078\% of images as compared to ImageNet-1k (1.28M images). We also see a significant difference between OFDBs and the ImageNet-1k$^\diamondsuit$ subset in terms of average accuracy. In the proposed methods, the configuration of 2D-OFDB-1k with random pattern augmentation has a better average rate in the table. Thereafter, in the experiments, we assigned random pattern augmentation for 2D-OFDB.

\noindent \textbf{Scaling experiments on ImageNet-1k (Table~\ref{tab:comparison_imagenet1k}).}
In order to investigate the scalability of OFDBs, we increased the number of categories from 1,000 to 21,000 and applied these categories to ViT-T and ViT-Base (ViT-B) in Table~\ref{tab:comparison_imagenet1k}.
For comparison, the results of ImageNet-21k, FractalDB-21k, ExFractalDB-21k, and RCDB-21k are reported.
\cRyo{ImageNet-21k consists of 14M images, and the other three FDSL datasets} consist of 21M images.
The results for the ImageNet-21k$^\diamondsuit$ subset involving one image per category are also reported.

With ViT-T, 2D-OFDB-21k outperforms FractalDB-21k (73.8 vs. 73.0) and is comparable with ExFractalDB-21k (73.7 vs. 73.6).
However, we see the performance gap between 2D-OFDB-21k and ImageNet-21k (73.8 vs. 74.1).
This is because the images of ImageNet-21k for pre-training and the images of ImageNet-1k for fine-tuning overlap.
\cRyoo{This gap is reversed in ViT-B, where 2D-OFDB-21k is recorded 0.4 points higher than ImageNet-21k on ImageNet-1k fine-tuning.}
% This gap was filled with ViT-B, where both ImageNet-21k and 2D-OFDB-21k were recorded as 81.8 on ImageNet-1k fine-tuning. 
Moreover, the 3D-OFDB-21k pre-trained model was recorded 0.9 point higher than the model pre-trained on ImageNet-21k.
Note that the computational \cRyoo{time} for pre-training in terms of GPU hours is reduced by \cHiro{78.7\%} (1,088 vs. 5,120 in terms of GPU hours). We clarified that ViT pre-training can be more efficient in terms of both data amount and computational time.

% ViT-Tでは、2D-OFDB-21kはFractalDB-21kを上回り（73.5 vs 73.0）、ExFractalDB-21kと同等（73.5 vs 73.6 ）である。
% しかし、2D-OFDB-21kとImageNet-21kの間には性能差があることがわかる（73.5対74.1）。

% このギャップはViT-Bで埋められ、ImageNet-21kと2D-OFDB-21kは共にImageNet-1kの微調整で81.8と記録された。

% さらに、3D-OFDB-21kで事前学習したモデルは、ImageNet-21kで事前学習したモデルよりも0.9ポイント高く記録されています。

% また、GPU時間換算で78.7%削減（GPU時間換算で1,088対5,120）されました。このように、ViTによる事前学習は、データ量と計算時間の両面において、より効率的な学習が可能であることが明らかとなりました。

\noindent \noindent \textbf{Training on small datasets (Table~\ref{tab:comparisons_vit_pretraining_limited_data}).}
Instance Discrimination with Multi-crop and CutMix
(IDMM), proposed by Cao \textit{et al.}~\cite{ViT2040}, which requires only 2,040 images for pre-training, is one of the most successful approaches for training ViTs on small datasets.
In Table~\ref{tab:comparisons_vit_pretraining_limited_data}, we compare OFDBs with IDMM.
Here, we used the fine-tuning configuration of IDMM for a fair comparison, {\it i.e.,} 800-epoch pre-training and 200-epoch fine-tuning on seven datasets of Flowers~\cite{Nilsback08_flowers}, Pets~\cite{pets}, DTD~\cite{dtd}, Indoor-67~\cite{indoor}, CUB~\cite{WahCUB_200_2011}, Aircraft~\cite{aircraft}, and Cars~\cite{Krause3DRR2013_cars}.
Note that IDMM has two settings: internal pre-training and external pre-training. We used PVT v2\cite{pvt_v2} for the ViT model, as in Cao \textit{et al.}
The former uses the same dataset for pre-training and fine-tuning. This performs well, as reported in~\cite{ViT2040}.
For the latter, we used an ImageNet subset, which consists of 2,040 randomly sampled images (with two or three images per category).
We refer to this as IDMM-ImageNet. 

The results are shown in Table~\ref{tab:comparisons_vit_pretraining_limited_data}.
Although we used a limited number of images to pre-train a ViT, the result is much higher than the accuracy of training from scratch using the same procedure of 200-epoch fine-tuning. The proposed method is also better than the other well-organized pre-training methods, 
%including IDMM~\cite{ViT2040}, SimCLR~\cite{ChenICML2020}, and SupCon~\cite{Peeters_2022}, with fewer pre-training images.
\cHiro{including IDMM~\cite{ViT2040} and SimCLR~\cite{ChenICML2020}, with fewer pre-training images.}
In comparison with IDMM-ImageNet, 2D-OFDB-1k pre-training still performs at a higher accuracy, indicating that 2D-OFDB-1k pre-training is highly \cHiro{beneficial}, even when using a limited number of \cHiro{synthesized images}.%artificially generated images without real images.

\noindent \textbf{\cHiro{COCO} detection/instance segmentation (Table~\ref{tab:comparison_detection_segmentation}).}
We validate object detection and instance segmentation on the
%the common objects in context (COCO)~\cite{LinECCV2014_coco} dataset.
\cHiro{COCO~\cite{LinECCV2014_coco} dataset.}
Here, we switch the backbone model from a ViT to a Swin Transformer~\cite{LiuICCV2021_Swin} with Mask R-CNN~\cite{HeICCV2017_mask} head. We perform training for 60 epochs on the COCO dataset. The proposed 2D-OFDB-21k pre-trained model scores are \cHiro{higher} than in the case of training from scratch and are similar to those for the model pre-trained with ImageNet-1k. \cHiro{2D/3D-OFDB-21k}
%, a one-instance FDSL of FractalDB-21k/ExFractalDB-21k, 
recorded similar rates of 67.1 and 64.4 (64.3 in 3D-OFDB-21k) at AP$_{50}$ in detection and segmentation tasks.

\tableG % COCO

\vspace{-3pt}
\subsection{Exploratory Study}
\vspace{-3pt}
% We explore the hyper-parameters in relation to a key parameter in OFDBs.
In this subsection, we basically apply the data augmentation methods and hyper-parameters used in the paper on the DeiT, unless we mention the changed parameters from those of the DeiT. We use fine-tuning accuracy on CIFAR-100 (C100) as an evaluation measure.

% \noindent \textbf{Batch size.} Table~\ref{tab:acc_and_batchsize} shows the relationship between the performance rate in OFDB pre-training and batch size. Since the used dataset contains 1,000 images in total, we evaluated batch sizes of \{64, 128, 256, 512\} in this experiment. We see that 256 is the best batch size for both 2D-OFDB and 3D-OFDB.

% \tableF % batch size

\noindent \textbf{Virtual camera for 3D-OFDB (Table~\ref{tab:oefdb_viewpoint})}. In ExFractalDB implementation, three axes (roll, pitch, yaw angle) are controlled to set the random viewpoint, which is then projected from the 3D model onto the 2D image.
Here, we project the 3D model onto 2D images with random viewpoints using \{1, 2, 3\} axes \{roll, pitch, yaw\}. Table~\ref{tab:oefdb_viewpoint} show the experimental results.
%According to the result from one axis with fixed (12) or free (360) viewpoints, the accuracy with 12 fixed viewpoints is better than the accuracy with free viewpoints. 
We use 12 fixed viewpoints at every 30 degrees in two or three axes.
Eventually, an improvement can no longer be expected by creating additional camera angles with roll, pitch, and yaw. Rather, a limited viewpoint from fixed 30-degree angles with only a yaw angle proves better.  %in 3D-OFDB pre-training.

% \tableE % Iteration

\begin{figure}[t]
 \centering
 \includegraphics[width=0.9\linewidth]{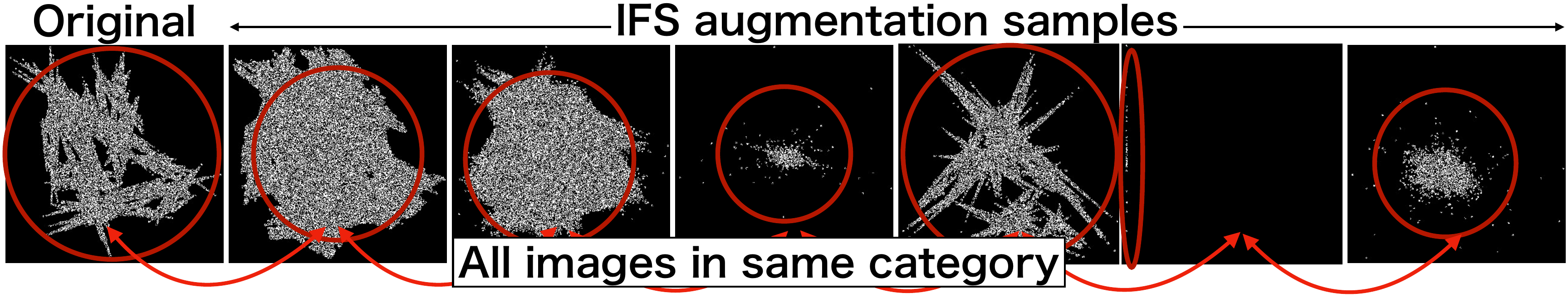}
 \vspace{-10pt}
 \caption{Sample images augmented with IFS}
 \label{ifs_samples}
 \vspace{-8pt}
\end{figure}

\noindent \textbf{Data augmentation (Table~\ref{tab:ofdb_dataaug})}.
Table~\ref{tab:ofdb_dataaug} shows the effects of data augmentation for fractal images.
We see that the proposed augmentation methods boost accuracy. In particular, random pattern augmentation is the most effective.
With IFS augmentation, we see the pre-training performance decrease significantly. 
This is because the better performance of 2D-OFDB-1k vs. FractalDB-1k in Tabel 1 is due to the removal of IFS augmentation.
Figure~\ref{ifs_samples} shows examples of image augmented with IFS. 
In the augmented images, there are images with significantly altered shapes, particularly images with almost meaningless shapes, which may hurt pre-training during DeiT augmentation.
% IFSで拡張されたサンプル例を図に示します．拡張された画像には，大きく形状を変化した画像や特に形状にほとんど意味のない画像が存在しでおり，DeiT拡張時には，これ等が事前学習に悪い影響を及ぼしていると考えられます．
Also, with rotation augmentation, which randomly rotates fractal images by 0, 90, 180, or 270 degrees, we see that the performance improvement is not significant. This is because rotation augmentation resembles flipping augmentation in the DeiT setting.

\tableH

\noindent \textbf{One-instance setting on real-image dataset \cHiro{(Table~\ref{tab:1ins_oin})}.} We conduct some additional experiments on the ImageNet-1k$^\diamondsuit$ subset. Here, we convert RGB images to gray-scale~\cite{gray_scale}, binary~\cite{oths_method}, and \cHiro{Canny} edge~\cite{canny} images. Table~\ref{tab:1ins_oin} shows the results of pre-training with these converted images. This table also shows the results for \cHiro{2D/3D-OFDB-1k}.

We see that binary and Canny images performed better than RGB images when we used these images in the pre-training phase. However, none of the images outperformed 2D/3D-OFDBs.
These results are consistent with the claim that ``\cHiro{object} contours are what matter in FDSL datasets'', as noted in a previous paper~\cite{Kataoka_2022_CVPR}. In fact, in order to pre-train a ViT model with one instance per category, an RGB representation is not enough on ImageNet-1k. A contour-emphasized dataset with a Canny edge detector is more efficient for pre-training a ViT, which is good at learning object contours. In the one-instance setting, we found that the pre-training effect with contour-emphasized image representations was also improved.

\tableI % augmentation

\begin{figure*}[t]
\hspace{-15pt}
% \vspace{-10pt}
 \scalebox{0.96}{
	\begin{tabular}{ccc}
    \begin{minipage}{0.33\textwidth}
		\centering
		\includegraphics[width=5.75cm]{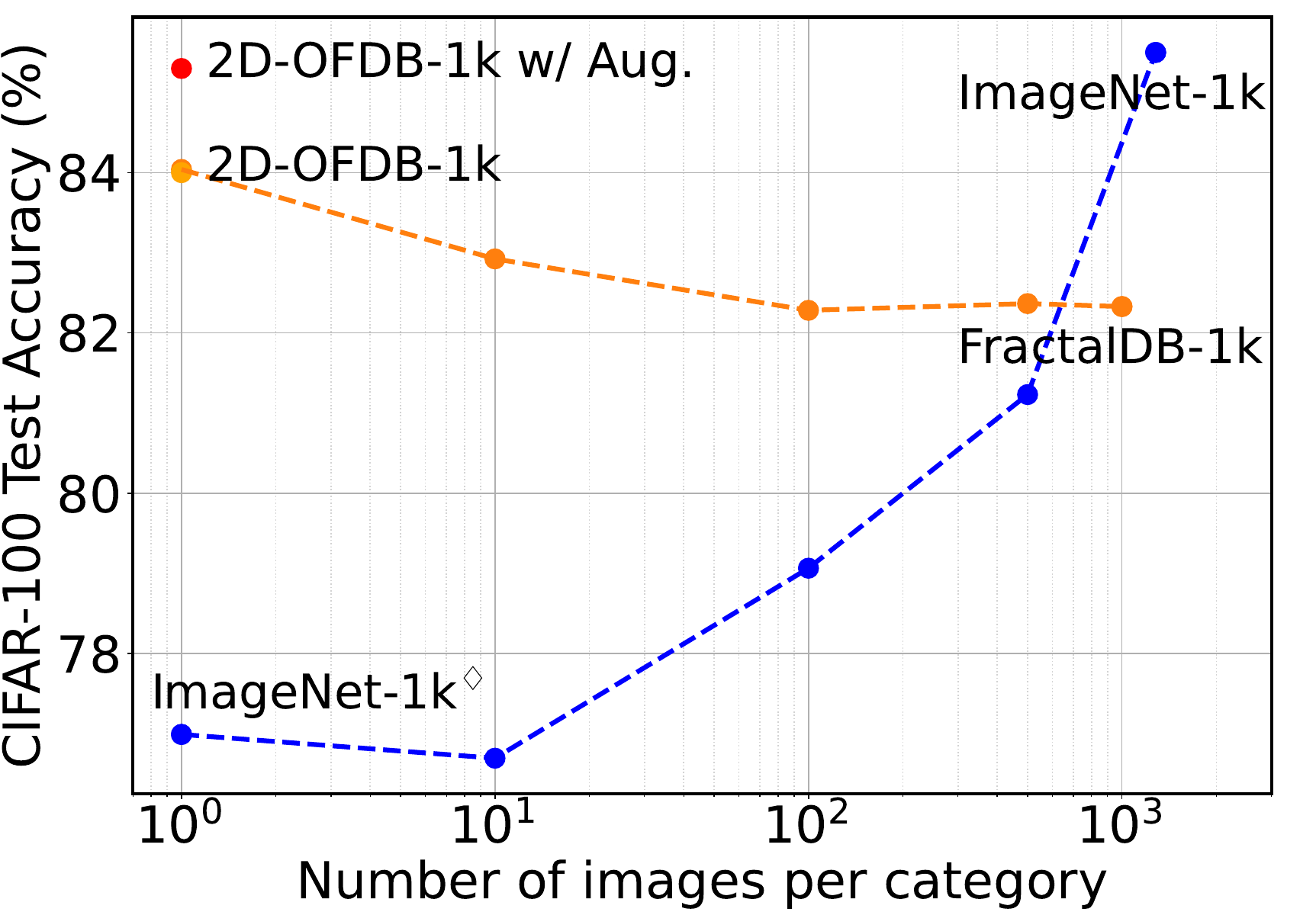}
		\vspace{-14pt}
		\caption{\cRyo{Relationship between accuracy and number of image instances per category.} %with a batch size of 256.
        }
		\label{fig:acc_and_instance}
	\end{minipage}
    \hspace{1pt}
	\begin{minipage}{0.33\textwidth}
		\centering
		\includegraphics[width=5.8cm]{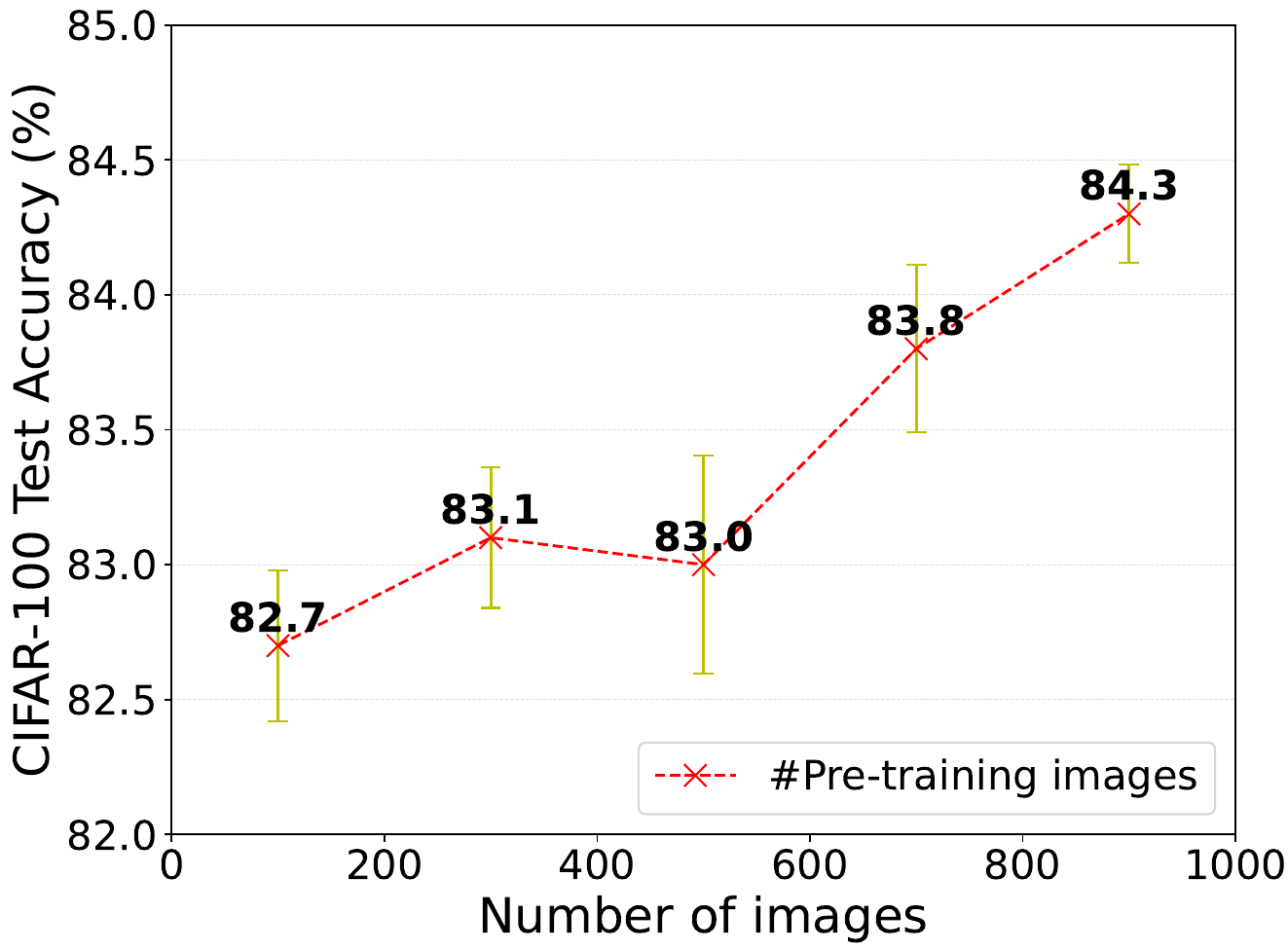}
		\vspace{-16pt}
		\caption{\cHiro{Relationship between number of pre-training images on OFDB and recognition accuracy.}}
		\label{fig:acc_numofimages_relationship}
	\end{minipage}
	\hspace{1pt}
    \begin{minipage}{0.33\textwidth}
		\centering
		\includegraphics[width=5.75cm]{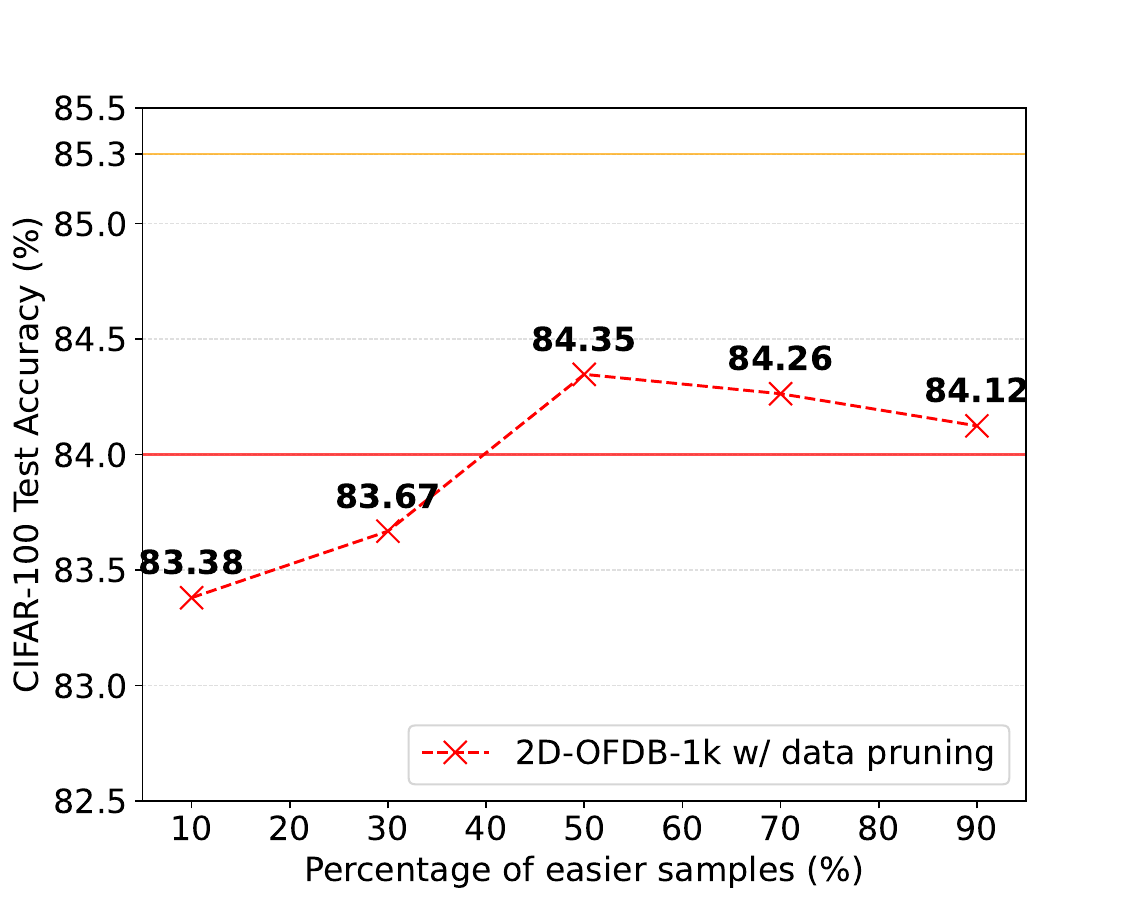}
		\vspace{-19pt}
		\caption{\cHiro{Data pruning for category selection on 2D-OFDB from 21k to 1k categories.}}
		\label{fig:datapruning}
	\end{minipage}
	\end{tabular}
 }
	\vspace{-18pt}
\end{figure*}

\noindent \textbf{Number of instances \cHiro{(Figure~\ref{fig:acc_and_instance})}.}
Figure~\ref{fig:acc_and_instance} shows the relationship between the number of image instances and accuracy in pre-training with ImageNet\cHiro{-1k} (SL) and 2D-OFDB-1k (FDSL). In the figure, the transition in accuracy is shown when \#instances are set to \{1, 10, 100, 500, 1,000\}. Formula-driven supervised learning has the highest accuracy when \#instances is 1; there is a slight decrease in accuracy up to 100 instances and almost no change thereafter. On the other hand, the lowest accuracy was observed for one instance when ImageNet-1k was used, and the accuracy improved with each increase in the number of instances. Note that since the total image dataset size is 1,000 images for the one-instance setting, the batch size is set to 256 for all settings. However, we assigned a better setting with a batch size of 1,024 on ImagNet-1k pre-training, except for the one-instance setting.
% XXはImageNetcHiro{-1k}で事前学習を行った際の画像インスタンス数と精度の関係を示す。(SL)と2D-OFDB-1k(FDSL)を用いた事前学習における画像インスタンス数と精度の関係を示しています。図では、インスタンス数を1、10、100、500、1,000に設定したときの精度の推移を示しています。数式駆動型教師あり学習は、「㊙」が1のときが最も精度が高く、100㊙までは若干精度が下がり、その後はほとんど変化がありません。一方、ImageNet-1kを用いた場合は、1インスタンスで最も精度が低く、インスタンス数が増えるごとに精度が向上することが確認されました。なお、1インスタンスの設定では、全画像データセットサイズが1,000画像であるため、バッチサイズはすべての設定で256に設定されています。しかし、ImagNet-1kの事前学習では、1インスタンス設定を除き、バッチサイズを1,024とし、より良い設定を割り当てました。

\cRyoo{For ImageNet, the performance of pre-training is reduced because the number of data is reduced, but for FractalDB, the performance is improved by reducing the number of data. The performance reduction can be attributed to the augmentation method of FractalDB's data. FractalDB augments the data with Table \ref{tab:ofdb_dataaug} Rotation and IFS to create a 1M-scale dataset. The results in Table~\ref{tab:ofdb_dataaug} show that Rotation does not improve the pre-training performance when using DeiT's data augmentation, and for IFS, it rather reduces the pre-training performance. Therefore, since reducing data in FractalDB corresponds to removing data augmented by IFS, we can consider that the accuracy of pre-training performance improves as the data decreases.}
% ImageNetは，データ数が減るため，事前学習の性能が落ちていると考察できるが，FractalDBにおいては，データ数が減ることが性能向上している．
% 性能が低下する理由としては、FractalDBのデータを増強する方法に原因があると考えられる。
% 性能が低下する理由としては，FractalDBのデータセットの作成ほうに要因がある．
% FractalDBでは，1M規模のデータセットを作るために，表xxのRotationやIFSでデータの水増しをおこなっている．
% 表xxの結果では，DeiTのデータ拡張の使用した場合，Rotationは事前学習効果に寄与せず，IFSについては，むしろ事前学習効果を悪化させていた．
% FractalDBにおいてデータを減らすということは，IFSによって生成されたデータを取り除くことに相当するため，FractalDBの推移は右肩下りになったと考えることができる．

\tableC

\noindent \cHiro{\textbf{Number of pre-training images on 2D-OFDB (Figure~\ref{fig:acc_numofimages_relationship}).}} We determined whether fewer pre-training images can work in ViT pre-training. Here, we decrease the number at \{100, 300, 500, 700, 900\} in the pre-training phase. Note that the number of parameter updates is aligned even if the number of images is decreased. And the results are then computed by averaging five times the pre-training effects of different 2D-OFDBs.Figure~\ref{fig:acc_numofimages_relationship} shows that the accuracy of pre-ViT training with only 100 synthesized images was 82.7 on C100, which is much higher than the accuracy for training from scratch (57.7) and still better than that for pre-trained on ImageNet with DINO supervision (82.4).

\noindent \cHiro{\textbf{Category selection with data pruning (Figure~\ref{fig:datapruning}).} We employed `accidentally' found fractal categories in FDSL pre-training. However, a one-instance setting in FDSL does not require augmented image instances; that is, it makes it easier to evaluate image categories on the FDSL dataset. Therefore, we tested whether category selection can improve the pre-training effects of OFDB with the data pruning method~\cite{neurips2022_ben_beyondscaling}.} 
% FDSLの事前学習では、「偶然」発見されたフラクタルカテゴリを採用している。しかし、FDSLにおける1インスタンス設定では、画像インスタンスを増やす必要がない、つまり、FDSLデータセット上で画像カテゴリを評価することが容易である。そこで、カテゴリ選択によりOFDBの事前学習効果を向上させることができるか、データ刈り込み手法により検証する~neurips2022_ben_beyondscaling}.incite{neurips2022_beyondscaling}.}. 

We analyzed the tendency of selected categories from the 21k to 1k category dataset. 
Figure~\ref{fig:datapruning} shows the relationship between easy sample usage with data pruning and fine-tuning accuracy. As the figure confirms, the balanced dataset (50:50 with easy:hard samples) recorded the best accuracy among all the settings. On the other hand, a dataset mainly consisting of hard samples (10:90 or 30:70 with easy:hard samples) had lower fine-tuning accuracy than the OFDB-1k pre-trained ViT-T. See supplementary material for the samples of selected categories with data pruning.
% 21kから1kのカテゴリーデータセットから、選択したカテゴリーの傾向を分析する。
% Figure~ref{fig:datapruning} は Easy sample usage with data pruning と Fine-tuning accuracy の関係を示している。図では、他の設定と比較して、バランスの取れたデータセット（簡単なサンプルと難しいサンプルが50：50）が最も良い精度を記録したことが説明されています。逆に、ハードサンプル中心のデータセット（イージー：ハードで10：90や30：70）は、OFDB-1kで学習したViT-Tよりも微調整の精度が低くなっています。データ刈り込みを行った選択カテゴリのサンプルは補足資料を参照。

% gmlpの結果
% \tableK
% \cRyo{
% \noindent \textbf{Performance on gMLP and ResNet.} Table~\ref{tab:model_type} shows the experimental results for gMLP with a 16$\times$16 patch ~\cite{LiuarXiv2021} and ResNet-50~\cite{HeCVPR2016} with 2D/3D-OFDB-1k. 
% %We employ gMLP-Tiny with a 16$\times$16 patch and ResNet-50. 
% The results show that ViT is more accurate than gMLP or ResNet in 2D/3D-OFDB-1k pre-training.
% }

\noindent \textbf{Processing time on dataset rendering.} Table~\ref{tab:rendering_time} shows a time comparison for dataset rendering in FDSL datasets. The table indicates that the proposed one-instance FDSL datasets, 2D/3D-OFDB-1k, are $\times$6.9/10.4 faster than the rendering time with FractalDB-1k and ExFractalDB-1k, respectively. By considering the rendering time, the speed increases are $\times$41.1/135.7 faster. Note that the number of 3D points in an image is much smaller than that of 2D rendering points. Therefore, the 3D-OFDB-1k is more efficient than 2D-OFDB-1k for total time.

% Table~ref{tab:rendering_time}は、FDSLデータセットにおけるデータセットレンダリングの時間比較を示しています。この表から、提案する1インスタンスFDSLデータセットである2D/3D-OFDB-1kは、FractalDB-1kとExFractalDB-1kによるレンダリング時間よりそれぞれ$6.9/10.4倍速いことが明らかである。また、レンダリング時間を考慮すると、$intimes$41.1/135.7高速化される。なお、画像中の3D点の数は2Dのレンダリング点の数よりはるかに少ない。したがって、3D-OFDB-1kは2D-OFDB-1kよりも総時間に関してより効率的である。
% 
% 

\begin{comment}
\begin{figure}[t]
\centering
\subfigure[\cHiro{Relationship number of pre-training images on OFDB and recognition accuracy.}]{\includegraphics[width=0.18\linewidth]{figure/ofdb_numof_images.pdf}
\label{fig:acc_numofimages_relationship}}
\subfigure[\cHiro{Data pruning for category selection on 2D-OFDB from 21k to 1k categories.}]{\includegraphics[width=0.18\linewidth]{figure/data_pruning_result.pdf}
\label{fig:numofimages_datapruning}}
\caption{Parameter tuning on RCDB. Tuning was conducted with \{C10, C100, Cars, Flowers\}. The values in the graphs show the average rates on the four datasets.}
\label{fig:additional_experiments}
\end{figure}
\end{comment}

% \begin{figure}[t]
%   \centering
%   \includegraphics[width=0.7\linewidth]{figure/ofdb_numof_imagesv2.pdf}
%   \vspace{0pt}
%   \caption{\cHiro{Relationship number of pre-training images on OFDB and recognition accuracy.}}
%   \vspace{0pt}
%   \label{fig:acc_numofimages_relationship}
% \end{figure}

\vspace{-5pt}
\section{Discussion and Conclusion}
\vspace{-3pt}

\noindent \textbf{ViT pre-training on a one-instance dataset.}  We validated ViT pre-training on an image dataset consisting of only one synthesized image per category. First, we succeeded in pre-training a ViT on the proposed OFDBs while still not requiring real images and human supervision. Moreover, the OFDB-1k pre-trained model performed better than the model pre-trained by FractalDB-1k (OFDB-1k 84.0 vs. FractalDB-1k 81.6; see Table~\ref{tab:comparison}). In further exploration, the proposed random patch augmentation with randomly selected 3$\times$3 pixels in the fractal image rendering proved to be the most effective way to improve the OFDB pre-training effect. Consequently, we added random patch augmentation to the OFDB (w/ aug 85.3 vs. w/o 84.0; see Table~\ref{tab:ofdb_dataaug}). We also confirmed that a synthesized image dataset containing one-instance per category is sufficient to pre-train a ViT model.

\noindent \textbf{Experimental comparison discussion.} The proposed OFDB-1k surpassed ImageNet-1k in pre-training for all tasks, including image classification, object detection, and instance segmentation (refer to Tables~\ref{tab:comparison} to \ref{tab:comparison_detection_segmentation}). Notably, 2D/3D-OFDB-21k matched or exceeded ImageNet-21k's pre-training performance. With ViT-B model fine-tuning, we achieved 82.2/82.7 accuracy on ImageNet-1k (Table~\ref{tab:comparison_imagenet1k}). Our pre-training utilized 21k images, compared to ImageNet-21k's 14M. The speedup between OFDB-21k and ImageNet-21k pre-training was about $\times$3.3 (2D/3D-OFDB-21k used 1,088 GPU hours vs. ImageNet-21k's 3,657 GPU hours). %in terms of GPU hours of pre-training.

% \実験的比較による考察 
% 実画像データセットとの比較により、提案するOFDB-21kは、画像分類、物体検出、インスタンス分割の全てのタスクにおいて、ImageNet-1k（1インスタンス）よりも優れた事前学習効果を示した（表～ref{tab:比較}, \ref{tab:comparison_imagenet1k}, \ref{tab:comparisons_vit_pretraining_limited_data}, および \ref{tab:comparing_detection_segmentation}を参照）。特に、提案した2D/3D-OFDB-21kは、事前学習効果においてImageNet-21kと同等かそれ以上であった。また、ViT-Bモデルを用いたImageNet-1kのfine-tuningでは、82.7を記録した（表～ref{tab:comparison_imagenet1k})。ImageNet-21kの14M画像に対して、21k画像で事前学習効果を得ることができました。また、OFDB-21kとImageNet-21kの事前学習では、約$3.3倍の高速化を達成しました（2D/3D-OFDB-21k 1,088 GPU hours vs. ImageNet-21k 3,657 GPU hours）。

In ViT pre-training with fewer images, 2D/3D-OFDB outperformed IDMM in average accuracy for benchmark fine-tuning datasets and pre-training image count (Table~\ref{tab:comparisons_vit_pretraining_limited_data}). Our method required no model modifications and fewer images, using just 1,000 synthetics versus IDMM's minimum of 2,040 reals. As shown in Figure~\ref{fig:acc_numofimages_relationship}, 2D-OFDB pre-training with 100 images yielded 82.7 accuracy on C100, surpassing DINO self-supervised ImageNet. This method substantially lowered image count in ViT pre-training compared to prior work.

\tableJ

% \vspace{-5pt}
\noindent \textbf{Limitations.} By using the proposed 2D/3D-OFDB pre-trained models, the pre-training methods still have lower accuracies compared to ImageNet-1k pre-training with a full-instance scale for relatively small datasets (see Table~\ref{tab:comparison}). We believe that when improved image representation and teacher labels are provided, even a 1,000-image dataset will achieve a superior pre-training effect than a 1.28M-image dataset. This would be advantageous for computing resources, especially in terms of memory usage and computational time. On the other hand, the models pre-trained on 2D/3D-OFDB-21k could not achieve the performance of ExFractalDB-21k pre-trained Swin Transformer (see Table~\ref{tab:comparison_detection_segmentation}). Though the detection/segmentation performance rates did not decrease greatly, we intend to explore whether this is a side-effect of reducing the number of instances and whether we can overcome this problem with other better image representations.

%  提案する2D/3D-OFDB事前学習モデルを用いることで、比較的小さなデータセットでは、フルインスタンススケールのImageNet-1k事前学習と比較して、事前学習法の精度がまだ低い（表~ref{tab:comparison}を参照）。我々は、改善された画像表現と教師ラベルが与えられた場合、1,000画像データセットでも128万画像データセットよりも良い事前学習効果を達成できると考えている。これは、計算機資源、特にメモリ使用量と計算時間の点で有利であると考えられる。一方、2D/3D-OFDB-21kで事前学習したモデルは、ExFractalDB-21kで事前学習したSwin Transformerの性能には及ばない（表~ref{tab:comparison_detection_segmentation}を参照）。検出・分割の性能は大きく低下していないが、これはインスタンス数を減らしたことによる副作用なのか、他の優れた画像表現でこの問題を克服できるのか、検討したい。

% \noindent \textbf{Future research.} We have succeeded in constructing cost-efficient pre-training datasets, namely, 2D/3D-OFDB. We implemented these datasets in terms of much less dataset construction, a small amount of memory usage, and a shorter pre-training speed with slightly higher pre-training effects through the proposed OFDBs. One potential next step would be to update the image representation while keeping in mind the important perspective of object contours in ViT pre-training. Hitherto, a pre-training method with FDSL has been expected to have a great potential to outperform pre-trained models with much larger datasets, such as JFT-300M/3B~\cite{SunICCV2017_jft300m} and IG-3.5B~\cite{MahajanECCV2018_ig3.5b}. 

\vspace{-5pt}
\section*{Acknowledgement}
\vspace{-3pt}
This paper is based on results obtained from a project, JPNP20006, commissioned by the New Energy and Industrial Technology Development Organization (NEDO). Computational resource of AI Bridging Cloud Infrastructure (ABCI) provided by National Institute of Advanced Industrial Science and Technology (AIST) was used.
We want to thank Junichi Tsujii, Yutaka Satoh, Mariko Isoagwa, Yasufumi Kawano, Ryosuke Yamada, Kodai Nakashima for their helpful comments in research discussions.

%%%%%%%%% REFERENCES
{\small
\bibliographystyle{ieee_fullname}
\bibliography{egbib}
}

\end{document}